\documentclass[letterpaper, 10 pt, conference]{ieeeconf}  

\IEEEoverridecommandlockouts                              

\usepackage{verbatim}
\usepackage{amsmath}

\usepackage[utf8]{inputenc}
\usepackage{graphicx}
\usepackage{subfigure}
\usepackage{float}
\usepackage{bm}
\usepackage{amssymb}
\usepackage{threeparttable}
\usepackage{multirow}
\usepackage{booktabs}
\usepackage{afterpage}

\usepackage{graphics} 
\usepackage{epsfig} 
\usepackage[T1]{fontenc}
\usepackage{color}

\usepackage[colorlinks=true]{hyperref}
\usepackage{cite}
\usepackage{booktabs} 
\usepackage{cleveref}

\graphicspath{{Images/}}

\title{\LARGE \bf
Safe and Individualized Motion Planning for Upper-limb Exoskeleton Robots Using Human Demonstration and Interactive Learning
}
\author{Yu Chen, Gong Chen, Jing Ye, Xiangjun Qiu, and Xiang Li
\thanks{Y. Chen, X. Qiu, and X. Li are with the Department of Automation, Tsinghua University, China. J. Ye and G. Chen are with the Shenzhen MileBot Robotics Co., Ltd, China. This work was supported in part by the Science and Technology Innovation 2030-Key Project under Grant 2021ZD0201404, in part by the National Natural Science Foundation of China under Grant U21A20517 and 52075290, and in part by the Shenzhen Science and
Technology Program under Grant KQTD20200909114235003. 
Corresponding author: Xiang Li (xiangli@tsinghua.edu.cn)}%
}

\begin{document}

\maketitle
\pagestyle{empty}  
\thispagestyle{empty} 

\begin{abstract}
A typical application of upper-limb exoskeleton robots is deployment in rehabilitation training, helping patients to regain manipulative abilities. However, as the patient is not always capable of following the robot, safety issues may arise during the training.
Due to the bias in different patients, an individualized scheme is also important to ensure that the robot suits the specific conditions (e.g., movement habits) of a patient, hence guaranteeing effectiveness. 
To fulfill this requirement, this paper proposes a new motion planning scheme for upper-limb exoskeleton robots, which drives the robot to provide customized, safe, and individualized assistance using both human demonstration and interactive learning. Specifically, the robot first learns from a group of healthy subjects to generate a reference motion trajectory via probabilistic movement primitives (ProMP). It then learns from the patient during the training process to further shape the trajectory inside a moving safe region. The interactive data is fed back into the ProMP iteratively to enhance the individualized features for as long as the training process continues. The robot tracks the individualized trajectory under a variable impedance model to realize the assistance.
Finally, the experimental results are presented in this paper to validate the proposed control scheme.

\end{abstract}

\section{Introduction}

In recent years, there has been escalating interest in the development and application of upper-limb exoskeletons. Numerous studies have demonstrated significant potential in a variety of domains\cite{Nef2006ARMinR}. 
Specifically, their applications have been extensively reported in the fields of power augmentation\cite{Kazerooni2008ExoskeletonsFH} and rehabilitation training\cite{Krebs1998RobotaidedN,Cui2017DesignOA,Zimmermann2023ANYexo2A}, where the latter is mainly to help patient regain manipulative abilities through repetitive movements.


Much progress has been made in improving the performance of upper-limb exoskeleton robots. 
On the hardware front, 
there has been a focused effort to overcome the limitations imposed by traditional rigid joint actuation mechanisms, leading to a call for innovative solutions \cite{Han2023HumanRobotIE}.
To address this challenge, the cable-driven mechanism, highlighted in \cite{Veneman2006ASE}, offers the advantages of reduced weight and increased flexibility. 
Similarly, the series elastic actuator (SEA), as explored in studies such as \cite{Veneman2007DesignAE} and \cite{Pan2018AdaptiveCB}, is another groundbreaking approach. 
These methods provide exoskeletons with the ability to minimize inertia, enhance backdrivability, and critically, to absorb excessive energy during sudden movements or impacts, thereby ensuring user safety.
From a software perspective, various control schemes have been developed under the paradigm of  ``assist as needed'' (AAN) \cite{Li2017AdaptiveHI}. In particular, 
Li et al. \cite{Li2018IterativeLI} introduced an iterative learning approach to eliminate disturbances during assistance. A multi-modal control strategy was proposed in \cite{Li2017MultimodalCS} to fulfill such a requirement, where the assistance is provided according to the interaction force between the human and the exoskeleton. 
Additionally, contemporary research has also explored the implementation of AAN by adaptively learning impedance parameters. 
This approach allows the exoskeleton to adjust its behavior based on real-time user feedback and interaction dynamics \cite{Han2023HumanRobotIE}.

As the patient closely interacts with the robot and may not always be capable of following its motion,
safety is the primary concern for the exoskeleton robot. Furthermore, the capability of providing individualized assistance can be treated as another important feature of the robot to ensure the training effect for multiple patients with significantly different medical conditions.  However, most existing upper-limb robots lack such features.

To address this, this paper proposes a new, safe, and individualized planning scheme for upper-limb exoskeleton robots through both offline and online learning in three steps.
\begin{enumerate}
    \item[1)] \emph{Offline Learning with Healthy Subjects}: The dataset of healthy human subjects is used train both a ProMP model and a safety grader, where the former is to parameterize upper-limb motion in consideration of motion redundancy and bias in different subjects, and the latter is to specify a safe region and vary the robot impedance.

    \item[2)] \emph{Offline Learning without Human Presence}: The robot is controlled to track trajectories in the dataset to approximate the unknown disturbance from the cable-driven mechanism, without human subjects wearing the robot, such that the disturbance can be isolated and then rejected in the form training process.

    \item[3)] \emph{Online Learning with Patients}: 
    A reference trajectory is generated by the ProMP model and then adjusted by assessing the patient's response using a safety grader. Individualizing the trajectory is realized when the patient's motion is kept inside the safe region while inhibiting abnormal behavior (probably due to the dysfunctionality of certain joints) with high robot impedance.
\end{enumerate}

The proposed method allows the exoskeleton robot to follow common human motion and tailor it by capturing the specific conditions and online response of the patient. 
The performance of the robot is validated in
a series of experiments and comparative studies by hiring healthy subjects to simulate disabled motion during rehabilitation training.

\section{System Structure}

\begin{figure}[!h]
    \centering
    \includegraphics[width=7cm]{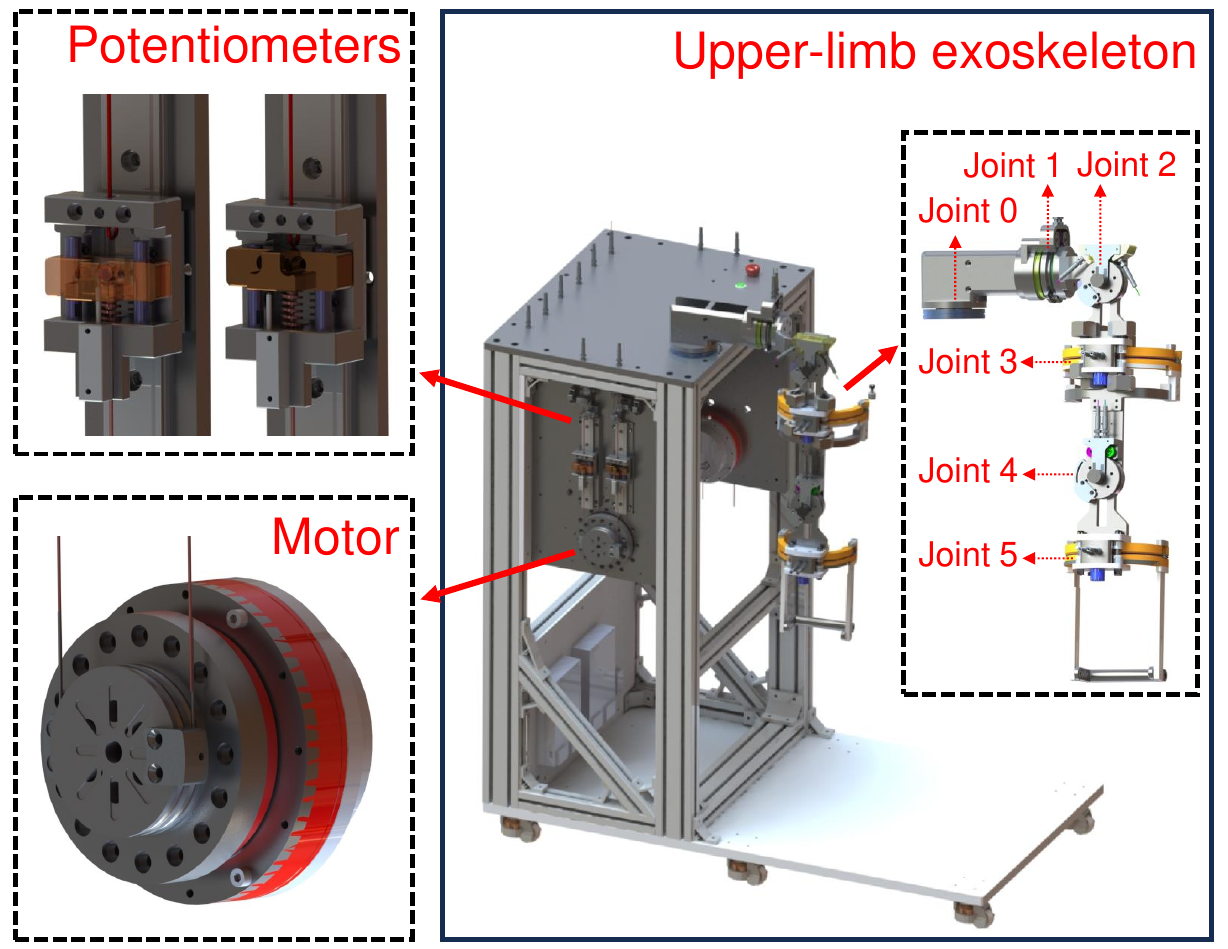}
    \caption{The structure of the cable-driven exoskeleton, comprising five active joints (Joints 1-5) and an additional passive joint(Joint 0).}
    \label{device}
    \vspace{-0.5cm}
\end{figure}

The architecture of our upper-limb exoskeleton is presented in Fig. \ref{device}, which comprises five active joints (i.e., Joint 1-5) and one passive joint (i.e., Joint 0). Note that the passive joint is specifically designed to coordinate the eccentric movement of the shoulder, and the other five joints are designed to achieve the following movements:
\begin{enumerate}
\item[-] Joint 1: Shoulder abduction/adduction 
\item[-] Joint 2: Shoulder flexion/extension 
\item[-] Joint 3: Upper-arm internal and external rotation
\item[-] Joint 4: Elbow flexion/extension
\item[-] Joint 5: Forearm internal and external rotation
\end{enumerate}

Each active joint is actuated with a SEA via the cable-driven mechanism to have the advantages of a light load and high impact tolerance. 
Each joint in this structure is equipped with an encoder (QY2204-SSI), a harmonic deceleration servo motor (AK80-64), and two potentiometers, as shown in Fig. \ref{device}.
Notably, the joints are actuated by motors through a linkage system involving cables and springs. 
This design facilitates the measurement of force inferred from the compression of springs that are integrated with the potentiometers.

\begin{figure}[!h]
    \centering
    \includegraphics[width=8.6cm]{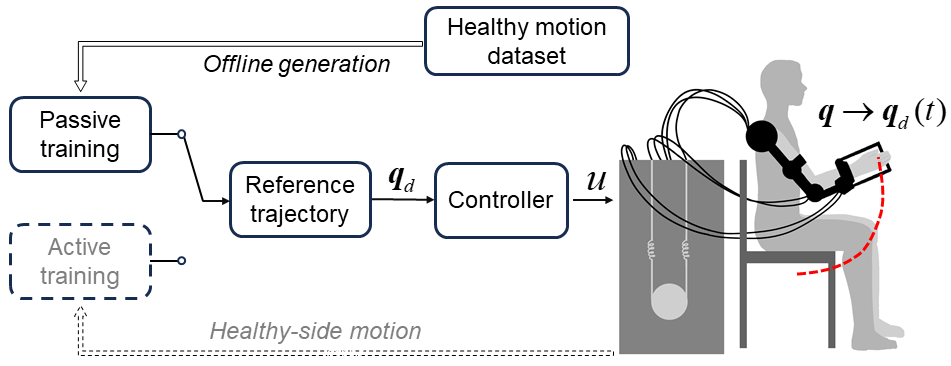}
    \vspace{-0.2cm}
    \caption{An illustration of the passive following mode for rehabilitation training, where a predefined trajectory $\bm q_d$ is generated in the offline phase and the robot is controlled to follow the trajectory (i.e., $\bm q$ goes to $\bm q_d(t)$) and lead the patient to carry out the training task.}
    \label{passive}
    \vspace{-0.4cm}
\end{figure}

Then, 
the dynamic model of such a robotic system (with both a cable-driven mechanism and SEA) can be described as
\begin{eqnarray}
&\bm M(\bm q)\ddot{\bm q}+\bm C(\dot{\bm q}, \bm q)\dot{\bm q}+\bm g(\bm q)=\bm K(\bm\theta-\bm q)+\bm\tau_e + \bm\tau_f,\label{dynRobot}\\
&\bm B\ddot{\bm\theta}+\bm K(\bm\theta-\bm q)=\bm u, \label{dynSEA}
\end{eqnarray}
where the notation in (\ref{dynRobot}) and (\ref{dynSEA}) are detailed in Table \ref{dynamic_table}. Specifically, $\bm\tau_o\hspace{-0.05cm}=\hspace{-0.05cm}\bm K(\bm\theta\hspace{-0.05cm}-\hspace{-0.05cm}\bm q)$ is the output torque of the SEA, connecting both the robot-joint subsystem (\ref{dynRobot}) and the motor-side subsystem (\ref{dynSEA}), and $\bm\tau_f$ represents unknown disturbance (e.g., friction) from the cable-driven mechanism. 
\begin{table}[h]
\caption{Dynamic parameters}
\centering
\begin{tabular}{c|l}
\hline
$\bm M(\bm q)\in\Re^{n\times n}$ & Inertia matrix of robot \\ \hline
$\bm C(\dot{\bm q}, \bm q)\in\Re^{n\times n}$ & Matrix related to centripetal and Coriolis forces \\ \hline
$\bm g(\bm q)\in\Re^n$ & Vector related to gravity \\ \hline
$\bm K\in\Re^{n\times n}$ & Stiffness matrix \\ \hline
$\bm q\in\Re^n$ & Robot joint angles\\ \hline
$\bm B\in\Re^{n\times n}$ & Inertia matrix of motor \\ \hline
$\bm \theta\in\Re^n$ & Motor rotation angles\\ \hline
$\bm\tau_e\in\Re^n$ & Interaction torque with human subject \\ \hline
$\bm\tau_f\in\Re^n$ & Disturbance torque \\ \hline
$\bm u\in\Re^n$ & Control input exerted on robot joints \\ \hline
\end{tabular}
\begin{tablenotes}
\small
\item[1] $n$ is the number of DoFs.
\end{tablenotes}
\label{dynamic_table}
\vspace{-0.6cm}
\end{table}
In general, the rehabilitation of the upper limb is either subject to an active mirroring mode or a passive following mode \cite{wang2011mirror,GonzlezMendoza2022DesignAI}. That is, the former directly maps the motion of the healthy side to that of the disabled side, while the latter drives the disabled side to follow a predefined trajectory (see Fig. \ref{passive}). This paper focuses on the passive training mode, which is applicable to patients whose both upper limbs lack the functional capabilities to execute daily activities. 
In such a mode:
\begin{enumerate}
    \item [1)] As both limbs of the patient are disabled, the reference trajectory is usually set by referring to other healthy human subjects, which are subject to sensor noises, motion uncertainty, and subject bias. 

    \item [2)] Assessing the patient's response and then adjusting the robot's motion is important to ensure the safety and effectiveness of the training. 

    \item [3)] The disturbance $\bm\tau_f$ due to the cable-driven mechanism is coupled with the interaction torque $\bm\tau_e$ in (\ref{dynRobot}), where the former should be rejected and the latter should be amplified. 
\end{enumerate}

This paper aims to overcome the above challenges and improve the safety and individualization (or effectiveness) of upper-limb rehabilitation.

\section{Motion Planning}
\begin{figure}[!h]
    \vspace{-0.5cm}
    \centering
    \includegraphics[width=1\linewidth]
    {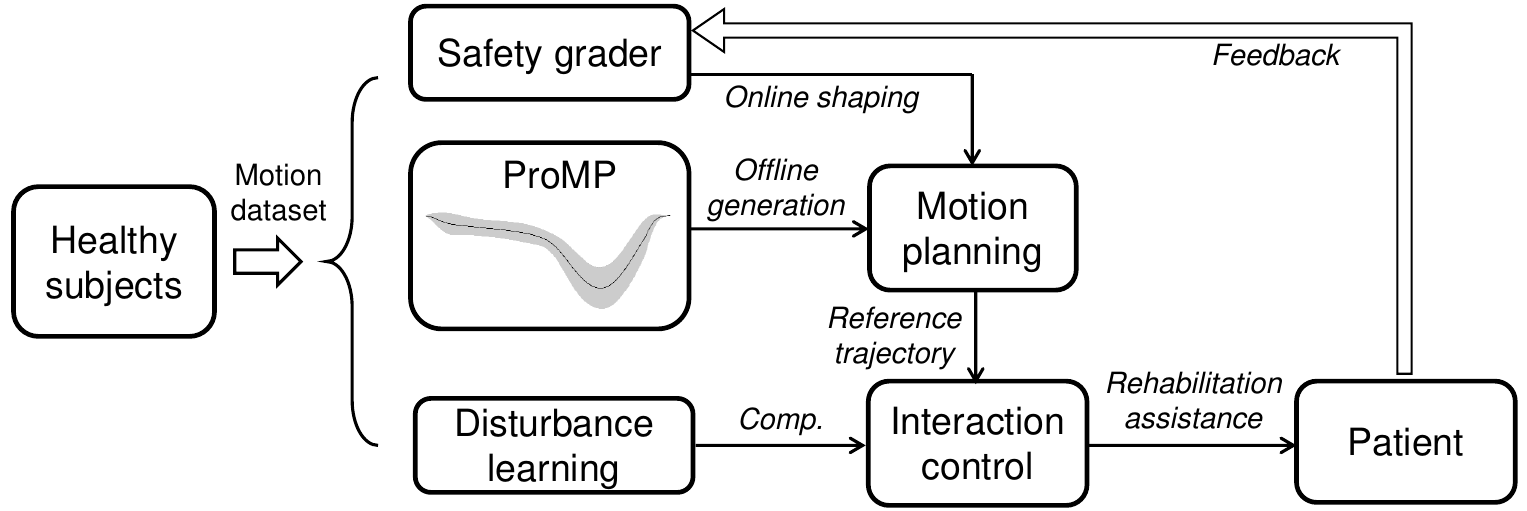}
    \caption{The workflow of the proposed method. Utilizing a motion dataset obtained from healthy subjects, the ProMP and safety grader are trained to offer offline trajectory generation and online trajectory refinement, respectively. Concurrently, the Disturbance Learning module compensates for friction arising during interactions. }
    \label{overall_frame}
    \vspace{-0.3cm}
\end{figure}

The structure of the proposed method is illustrated in Fig. \ref{overall_frame}. First, a motion dataset is collected from a group of healthy subjects who are asked to perform daily motion repetitively. Second, the dataset is used to train three modules in parallel: (i) {\em ProMP}, to generate a reference trajectory of rehabilitation training for the patient; (ii) {\em safety grader}, to adjust the trajectory online according to the feedback of the patient; (iii) {\em disturbance learning}, to compensate the unknown disturbance introduced by the cable-driven mechanism. 
As such, the robot not only learns from healthy subjects prior to the rehabilitation training, it also learns from the patient via interaction during the training. Such a scheme offers several benefits: it individualizes assistance,  regulates uncertainties, and deals with safety issues, all without interrupting the training process.
\\

\noindent\textbf{Offline Generation}: 
%
ProMP is a technique that encodes a collection of trajectories into a probabilistic model \cite{paraschos2013probabilistic}.
This model is capable of generating similar references through sampling.
Utilizing ProMP for sampling trajectories is particularly advantageous for repetitive movement in the passive training mode as it inherently accounts for sensor noise, human uncertainty, and individual biases.

To implement the ProMP, we express the trajectory by means of the weight vector $\bm \omega \in \Re^{Dn\times1}$, where $D$ is the number of basis functions, such that
\begin{align}
 \bm{y}_t &=  \left[ 
\begin{array}{ccc}
\bm q_{1,t}^T & \cdots & \bm q_{n,t}^T
\end{array} 
\right]^T=\bm\Psi_t^T \bm\omega+\bm\epsilon_y,\\
\bm q_{i,t} &= \left[ 
\begin{array}{cc}
q_{i,t}  & \dot{q}_{i,t}
\end{array} 
\right]^T,\\
p(\bm\tau_y|\bm w) &= \prod_t\mathcal{N}(\bm{y}_t|\bm\Psi_t^T \bm \omega, \bm \Sigma_y),
\end{align}
where $\bm q_{i,t}\in\Re^{2}$ stands for the composite vector of the $i^{th}$ joint at time step $t$, $\bm\epsilon_y\sim \mathcal{N}(\bm 0,\bm \Sigma_y)$ represents zero-mean i.i.d. Gaussian noise, $\bm\tau_y$ denotes the trajectory over the demonstration, and $\bm\Psi_t \in \Re^{Dn\times 2n}$, chosen as a Gaussian form\cite{paraschos2013probabilistic}, is the time-variant basis matrix.


Now, with the assumption of $\bm\omega \sim \mathcal{N}(\bm\omega|\bm \mu_{\omega}^{(k)},\bm \Sigma_{\omega}^{(k)})$, a new trajectory at time step $t$ can be modeled as
\begin{align}
p(\bm y_t;\bm \mu_{\omega}^{(k)},\bm \Sigma_{\omega}^{(k)}) = \int\mathcal{N}(\bm{y}_t|\bm\Psi_t^T \bm \omega, \bm \Sigma_y)\mathcal{N}(\bm\omega|\bm \mu_{\omega}^{(k)},\bm \Sigma_{\omega}^{(k)})d\bm\omega.\label{proModel}
\end{align}
Therefore, the desired trajectory is given as
\begin{align}
\bm q_d(t) = [q_{i,t},\cdots, q_{n,t},]
\end{align}
The parameters $\bm \mu_{\omega}^{(k)}$ and $\bm \Sigma_{\omega}^{(k)}$ can be deduced from the collected $k$ trajectories via the expectation maximization algorithm \cite{lazaric2010bayesian}, thereby constructing the probabilistic model.\\

\begin{figure}[!h]
    \centering
    \includegraphics[width=7.5cm]{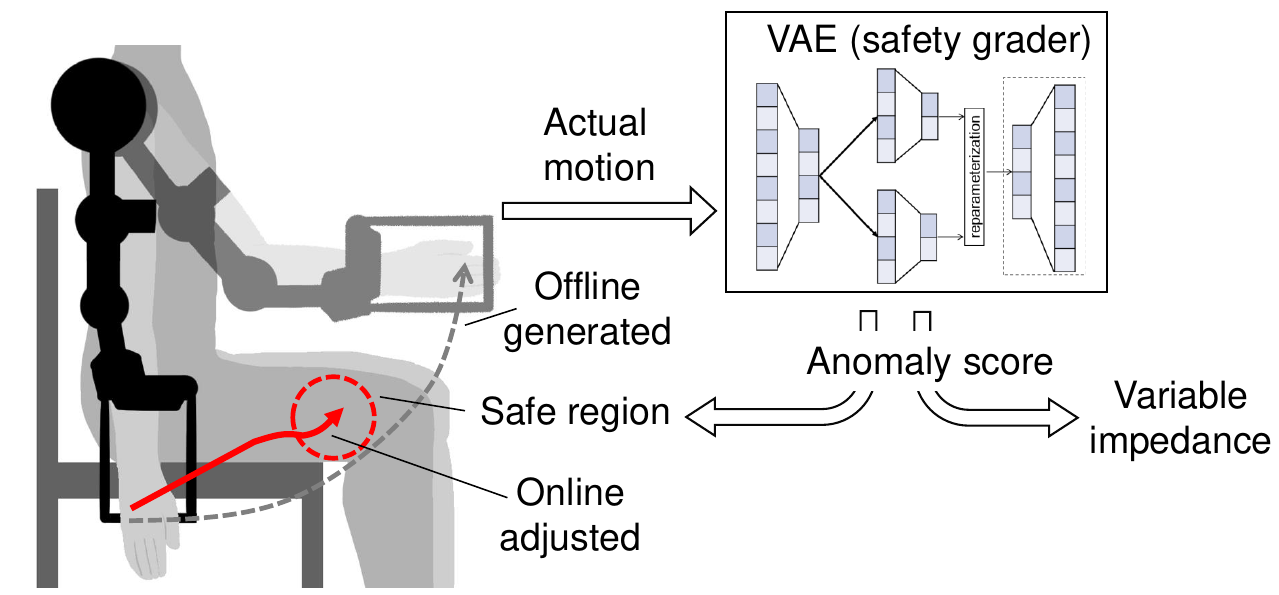}
    \caption{The VAE is trained based on the healthy motion database and treated as the safety grader. The anomaly score outputted by the grader is used to form a safe region so that the adjusted trajectory stays inside it to ensure the patient can safely follow the trajectory. The anomaly score is also used in the variable impedance model. }
    \label{plan_demo}
    \vspace{-0.3cm}
\end{figure}

\noindent\textbf{Online Adjustment}: Due to disabled functionality, the patient may be unable to fully follow the trajectory based on healthy subjects, likely for specific joint configuration during the training process. To address any potential conflicts, a safety grader is designed to further shape the generated trajectory according to the actual motion of the patient. As seen in Fig. \ref{plan_demo}, the adjusted trajectory may deviate from the planned one, but will always be within a dynamic safety region, ensuring safe interaction between the robot and patient.

Based on the collected movement dataset, a variational autoencoder (VAE), serving as a grader, is trained to assess the uncertainty associated with a given trajectory
\begin{align}
s = f(\bm q,\bm \tau_o),
\end{align}
In this context, $s$ stands for the online uncertainty evaluation score, indicating the deviation between the present trajectory and interaction force relative to the baseline observed in healthy cases.
The safe region, indicating low uncertainty (i.e., low score), is further depicted in Fig. \ref{Weighting_fig}.
Given a discrete sample time $\Delta t$, the update of the score is obtained via the linearization process:
\begin{align}
s^{k+1} = s^k +  \left[ (\frac{\partial f}{\partial \bm q})^T \dot{\bm q} + (\frac{\partial f}{\partial \bm \tau_o})^T \dot{\bm \tau}_o\right]\Delta t
\label{s_dynamic}
\end{align}

To adjust the trajectory according to the evaluation score, we identify the state vector as $\bm x = [s, \bm q_d^T, \dot{\bm q}_d^T]^T$ and further define the acceleration as $\bm u_a$. 
Subsequently, we incorporate (\ref{s_dynamic}) into the planning framework to yield a composite kinematic model:
\begin{align}
\bm x^{k+1} &= \bm A\bm x^{k} + \bm B\bm u_a^{k},
\end{align}
where 
\begin{align}
\bm A&= 
\left[ 
\begin{array}{ccc}
1  & \bm 0 & (\frac{\partial f}{\partial \bm q})^T\Delta t \\
\bm 0 & \bm I  & \bm I \Delta t \\
\bm 0 & \bm 0 & \bm I
\end{array} 
\right] \\
\bm B&= 
\left[ 
\begin{array}{c}
0 \\
\bm 0 \\
\bm 1 \Delta t
\end{array} 
\right]
\end{align}
and $\bm I \in \Re^{n\times n}$ defines the identity matrix, $\bm 1 \in \Re^{n \times1}$ denotes a vector with all elements set to $1$.

By employing the model predictive control (MPC), the cost to be minimized is specified as
\begin{align}
 \mathcal{L}(\bm u_a) = & \sum_{t}^{t+N_p}(\|\bm x_i - \bm x_{ri}\|_{\bm Q}^2 + \|\bm u_{ai}\|_{\bm R}^2),
 \label{cost_function}
\end{align}
where $\bm x_{ri}=[0, \bm y_i^T]^T$ is the reference vector, $N_p$ refers to the predictive horizon, subscript $i$ signifies the time step $i$, and $\bm Q\hspace{-0.05cm}\in\hspace{-0.05cm}\Re^{(2n+1)\times (2n+1)}, \bm R\hspace{-0.05cm}\in\hspace{-0.05cm}\Re^{n\times n}$ are symmetric positive-definite weighting matrices. Minimizing  (\ref{cost_function}) is done to ensure that the planned trajectory remains within the bounds of the safe angle range and velocity.
Note that this process is a closed loop, where $\bm x_t$ is updated in each time step $t$.

After the adjusted trajectory is implemented in the exoskeleton robot to assist the patient, the patient's actual motion during training is also recorded to iteratively update the
probabilistic model (\ref{proModel}). That is,
$\bm \mu_{\omega}^{(k)} \leftarrow \bm \mu_{\omega}^{(k+1)}$ and $\bm \Sigma_{\omega}^{(k)} \leftarrow \bm \Sigma_{\omega}^{(k+1)}$, which is then used
to facilitate the planning of subsequent trajectories. With the injection of motion data from the specific patient, the further generated trajectory would become increasingly individualized, hence better suiting the patient’s specific medical conditions.
Therefore, with the proposed method, the exoskeleton robot can explore online interaction to improve both interaction safety and training effects. 

\section{Interaction Control}
\label{Interaction_controller}
In this section, an interaction control method is proposed to achieve two objectives: (i) approximate then compensate the disturbance torque, and (ii) drive the robot to follow the adjusted trajectory under a variable impedance model for rehabilitation training.\\

\noindent\textbf{Disturbance Rejection}: 
As the disturbance torque $\bm\tau_f$ is coupled with the interaction torque $\bm\tau_e$ in (\ref{dynRobot}), it is not trivial to isolate the disturbance without any external sensor. However, such disturbances should be properly compensated to improve tracking accuracy.  

To achieve this, the robot first works without the presence of the wearer, where $\bm\tau_e\hspace{-0.05cm}=\hspace{-0.05cm}\bm 0$. Then, note that
the disturbance torque can be parameterized as \cite{de1995new}
\begin{align}
\bm\tau_f &= (\bm a_f + \bm b_f \odot e^{-\bm c_f \odot\dot{\bm q}} + \bm d_f\odot\dot{\bm q})\odot\bm {sgn}(\dot{\bm q})\notag \\
& \approx (\bar{\bm a}_f + \bar{\bm b}_f\odot\dot{\bm q} + \bar{\bm c}_f\odot\dot{\bm q}\odot\dot{\bm q})\odot\bm {sgn}(\dot{\bm q})= \bm Y(\dot{\bm q})\bm \psi,\label{friction_model}
\end{align}
where $\bm a_f, \bm b_f, \bm c_f, \bm d_f$ are the unknown parameters, $\bar{\bm a}_f, \bar{\bm b}_f, \bar{\bm c}_f$ are derived from the Taylor expansion as simplifications for the model,
$\odot$ denotes the Kronecker product, $\bm {sgn}(\cdot)$ represents a sign function, $\bm Y(\cdot)$ refers to a regressor matrix, and $\bm\psi$ symbolizes the vector of model parameters. 
Note that the approximation presented in (\ref{friction_model}) is reasonable because the velocity of the robot joints remains relatively low during the rehabilitation process.


In the presence of unknown disturbance, the vector $\bm\psi$ is unknown and estimated with an online updated law as
\begin{align}
  & \dot{\hat{\bm \psi}} = \bm \Gamma \bm Y^{T}(\dot{\bm q})[\dot{\bm q}-\dot{\bm q}_f+\alpha (\bm q-\bm q_f))],
  \label{reg_f}
\end{align}
where $\hat{\bm \psi}$ is the estimate of $\bm\psi$, $\bm \Gamma$ is a positive definite scaling diagonal matrix, $\alpha$ is a positive constant, and $\bm q_f$ is a time-varying trajectory which is set to be persistently excited (P.E.) \cite{arimoto1996control}. Note that $\bm q_f$ is not related to the rehabilitation but is used to approximate the disturbance. 
Next, it can be proved that the updated law leads to the convergence of $\hat{\bm\psi}\hspace{-0.1cm}\rightarrow \hspace{-0.1cm}\bm\psi$ \cite{arimoto1996control}, when a control scheme (e.g., backstepping approach \cite{saberi1990global,pan2017adaptive}) is designed for the trajectory-tracking task. Hence, the disturbance will be well approximated in the steady state and denoted as $\hat{\bm\tau}_f$.\\

\begin{figure}[!h]
    \centering
    \includegraphics[width=1\linewidth]{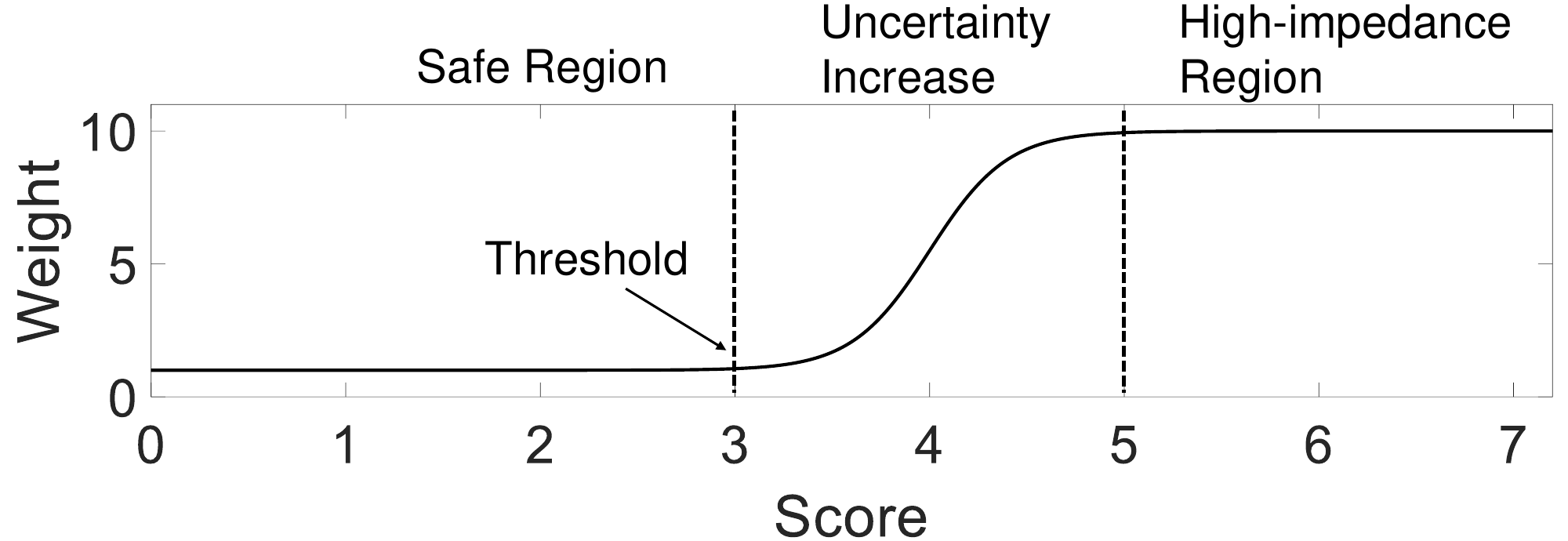}
    \caption{An example of the weighting function with impedance modulation triggered by the score exceeding the safe region. Specifically, $\lambda_1=-4.5,\lambda_2=5.5,m = 0.4,h=10$.
    Note that the "Uncertainty Increase" zone is demarcated by $3< s < 5$, within which the robot modulates its impedance to accommodate a lower tracking error.}
    \label{Weighting_fig}
    \vspace{-0.3cm}
\end{figure}

\noindent\textbf{Variable Impedance}:
The variable impedance model is introduced to regulate the interaction between the robot and the patient as \cite{zhang2023multi}
\begin{eqnarray}
&\bm C_d(\dot{\bm q}-\dot{\bm q}_d)+\bm K_d(\bm q-\bm q_d)=\frac{1}{w(s)}\bm\tau_e.\label{impedancemodel}
\end{eqnarray}
where $\bm C_d, \bm K_d\in\Re^{n\times n}$ are diagonal and positive-definite matrices, specifying the impedance parameters, $\bm q_d$ is the reference trajectory subject to both offline generation and online adjustment (introduced in Section III), and $w(\cdot)$ is a weighting function of the evaluation score as \cite{zhang2023multi}
\begin{eqnarray}
&w(s) = \lambda_1 \tanh(-\frac{s}{m} + h) +\lambda_2 \label{weight_fun}
\end{eqnarray}
where $\lambda_1$ and $\lambda_2$ are two positive constants that determine the range and midpoint of the weighting function. The constant $m$ normalizes the anomaly score into a specific, confined range, while $h$ signifies the offset 
from the origin along the positive direction of the horizontal axis.
An example of the weighting function is shown in Fig. \ref{Weighting_fig}, in which the safe region is defined by a threshold.

\begin{figure*}[!t]
\vspace{-0.5cm}
\centering
\subfigure[]{
    \label{joint3}
    \includegraphics[width=2.1in]{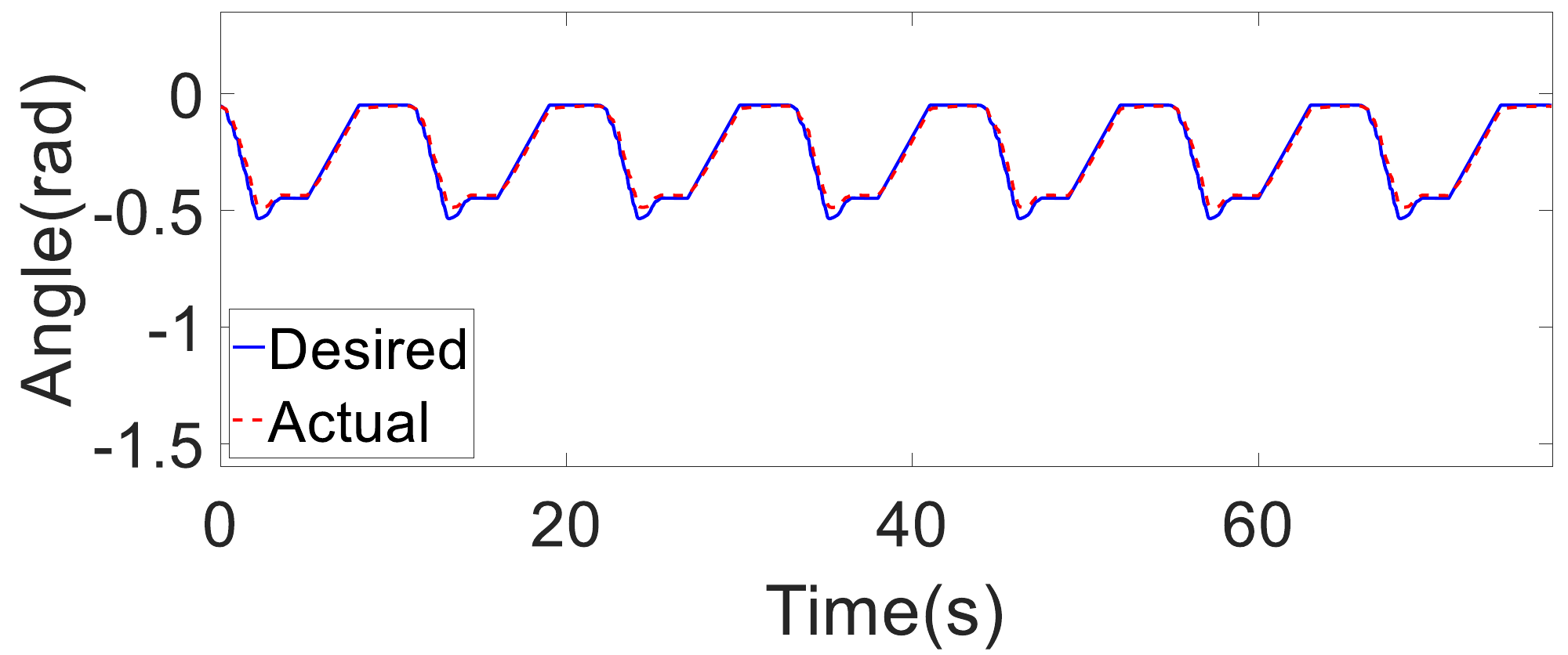}}
\subfigure[]{
    \label{joint4}
    \includegraphics[width=2.1in]{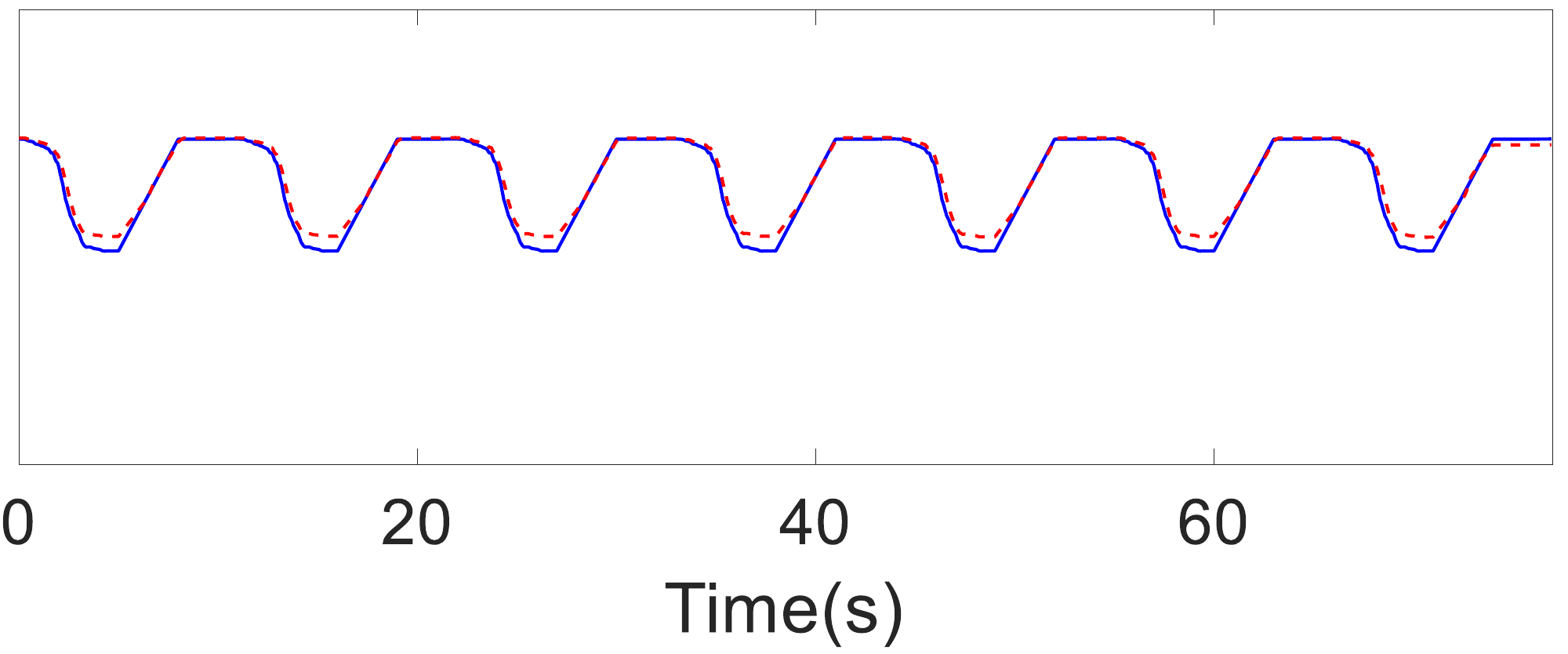}}
\subfigure[]{
    \label{joint5}
    \includegraphics[width=2.1in]{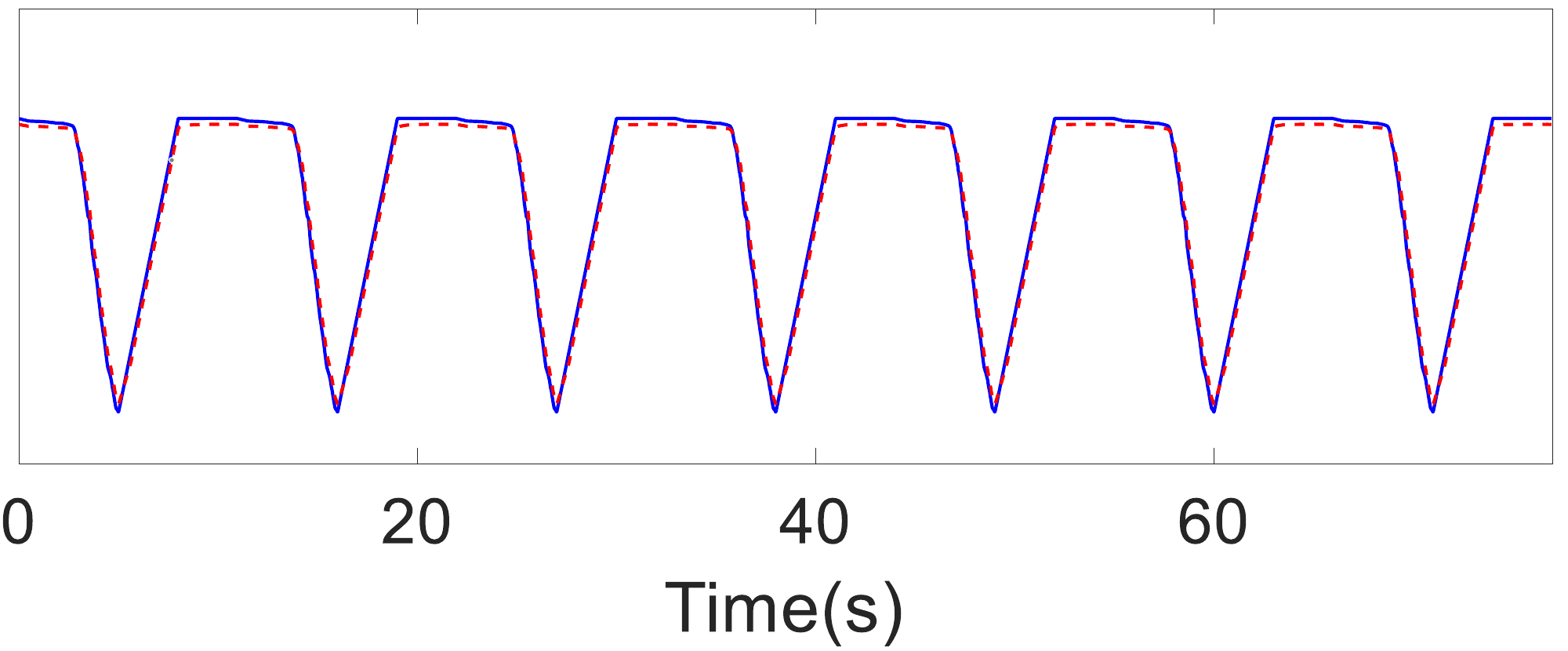}}
\caption{Trajectory-tracking control with disturbance compensation: (a) Joint 1, (b) Joint 2, (c) Joint 4. The blue solid line and red dashed line represent the desired and actual trajectory, respectively.}\label{track}
\vspace{-0.4cm}
\end{figure*}

\begin{figure*}[!t]
\centering
\subfigure[]{
    \label{joint3}
    \includegraphics[width=2.1in]{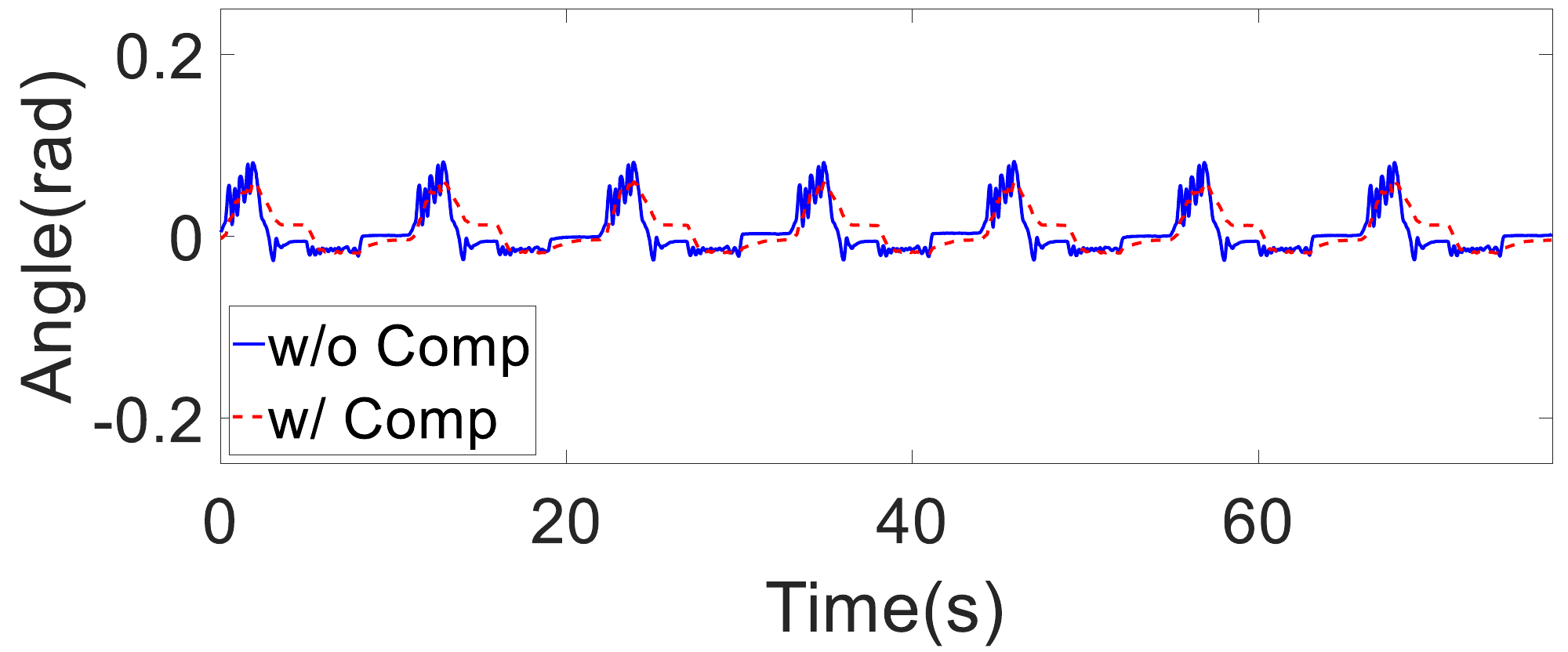}}
\subfigure[]{
    \label{joint4}
    \includegraphics[width=2.1in]{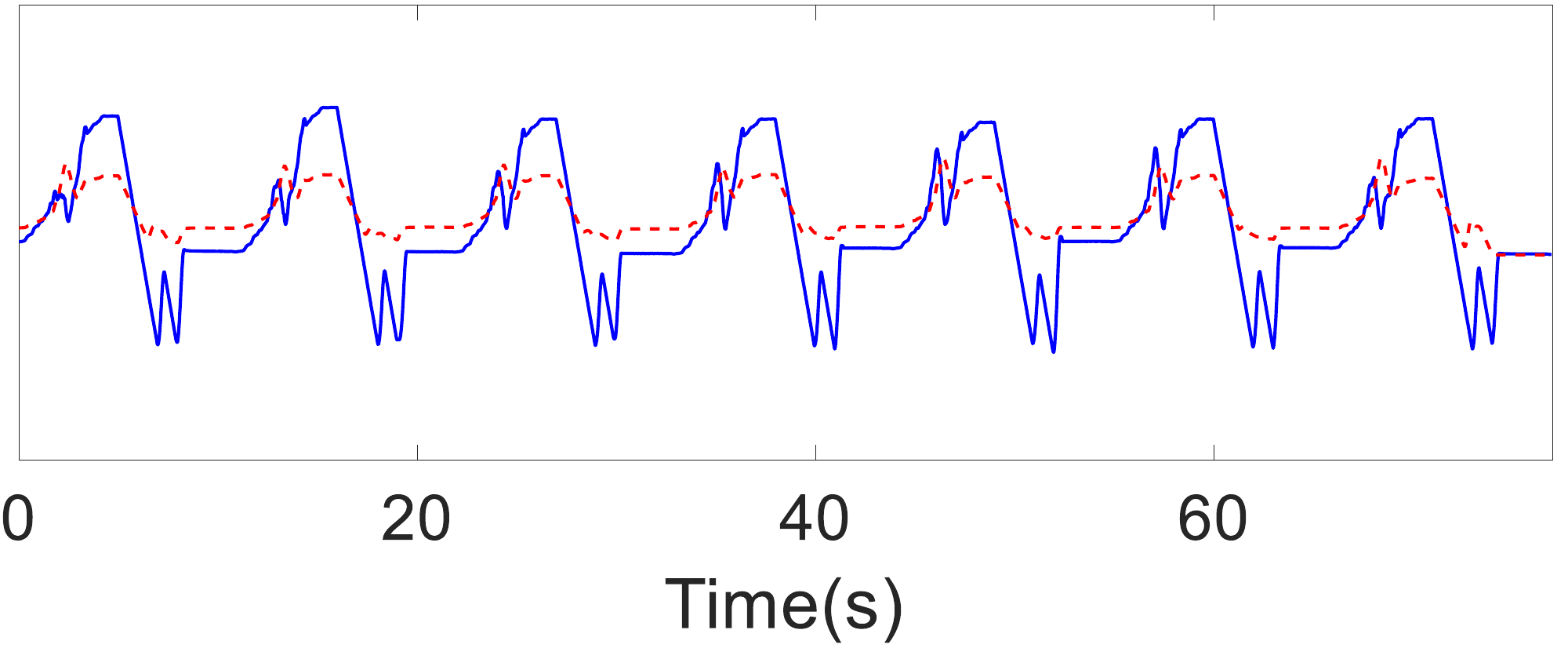}}
\subfigure[]{
    \label{joint5}
    \includegraphics[width=2.1in]{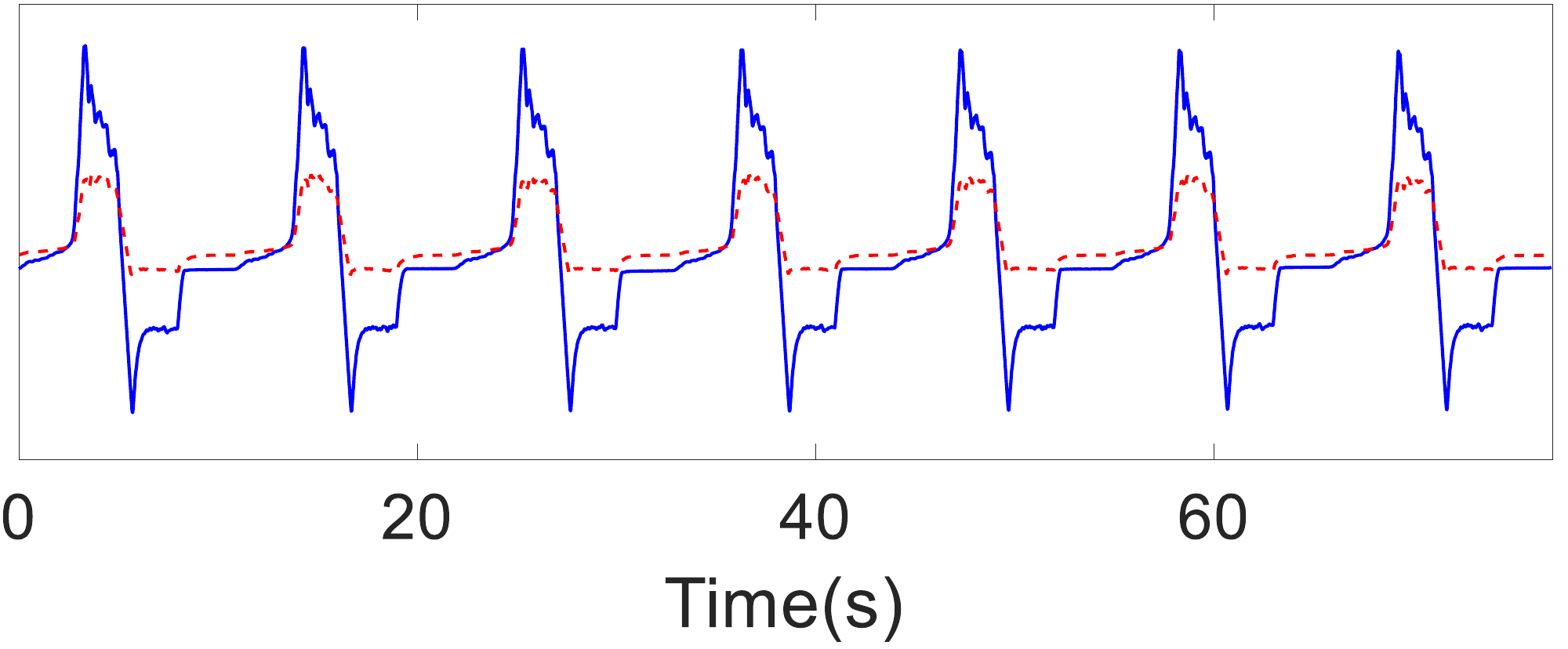}}
\caption{Comparative studies on tracking control with or without disturbance compensation: (a) Joint 1, (b) Joint 2, (c) Joint 4.
The blue solid line and red dashed line represent the results without and with disturbance compensation, respectively.}\label{trackerror}
\vspace{-0.4cm}
\end{figure*}

Using such a variable impedance model (\ref{impedancemodel}) can improve safety in three ways:
\begin{enumerate}
    \item [1)] Compared with purely position control, it allows the patient to deviate from the reference trajectory. 

    \item [2)] The individualized trajectory matches the specific condition of the patient, and such a feature will be amplified as the rehabilitation training continues. 

    \item [3)] The robot becomes more rigid by increasing the impedance to keep the patient's actual motion always inside the safe region.
\end{enumerate}

Then, the variable impedance model is represented with an impedance vector as \cite{cheah1998learning}
\begin{align}
\bm z&=\dot{\bm q}-\dot{\bm q}_r\nonumber \notag \\
&=\dot{\bm q}-\dot{\bm q}_d+\bm C_d^{-1}\bm K_d(\bm q-\bm q_d)-\frac{1}{w(s)}\bm C_d^{-1}\bm\tau_e,\label{vectorz}
\end{align}
where $\dot{\bm q}_r=\dot{\bm q}_d-\bm C_d^{-1}\bm K_d(\bm q-\bm q_d)+\frac{1}{w(s)}\bm C_d^{-1}\bm\tau_e$
is defined as a reference vector.

Now, the interaction control input can be given as
\begin{align}
    \bm u&= - \bm K_v(\dot{\bm q}-\dot{\bm\theta}) - \hat{\bm\tau}_f -\hat{\bm\tau}_e \nonumber\\
    &-\bm K_z \bm z+(\bm M(\bm q)+\bm B)\ddot{\bm q}_r+\bm C(\dot{\bm q}, \bm q)\dot{\bm q}_r
+\bm g(\bm q),\label{overall_control_terms}
\end{align}
where $\bm K_v, \bm K_z$ are diagonal and positive-definite matrices and $\hat{\bm\tau}_e$ is the estimate of $\bm\tau_e$ and obtained with a model-based observer \cite{mohammadi2017nonlinear}. In (\ref{overall_control_terms}), the first term is to stabilize the actuator subsystem, the second term is to reject the disturbance, the third and the fourth terms are to realize the variable impedance model, and the remaining terms denote the dynamic compensation. It can be proved by referring to the singular-perturbation theory \cite{shu2023twostage} that $\bm z\rightarrow \bm 0$ as $t\hspace{-0.05cm}\rightarrow\hspace{-0.05cm}\infty$, hence the robot is controlled to help the patient carry out rehabilitation training under the variable impedance model.

\section{Experiment}
\begin{figure}[!h]
    \vspace{-0.2cm}
    \centering
    \includegraphics[width=1\linewidth]{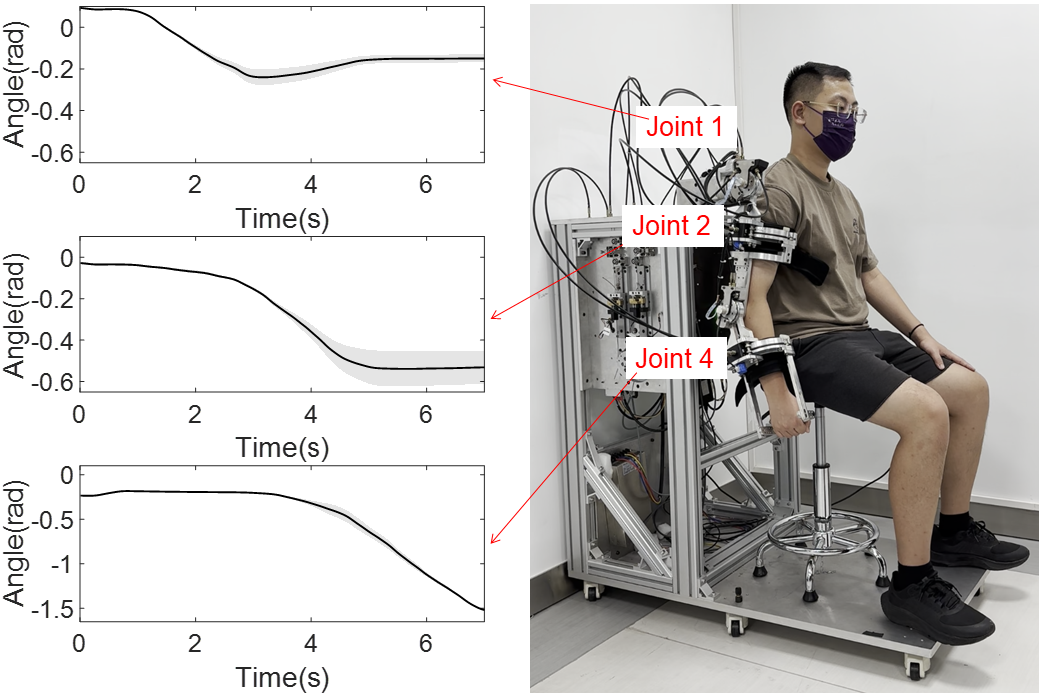}
    \caption{A human subject wears the upper-limb exoskeleton robot.
    The left column shows the results of ProMP modeling in the chosen active joints.
    The black solid line represents the mean of the trajectories, with the surrounding grey shade denoting its standard variation.}
    \label{real_device}
    \vspace{-0.5cm}
\end{figure}

Experiments with the developed exoskeleton robot (see Section II) have been performed to validate the proposed control scheme. Specifically, the proposed method was implemented in three robot joints (Joint 1, 2, 4), as shown in Fig. \ref{real_device}. 
Owing to the absence of authorization for clinical trials, we conducted our experiments on healthy individuals to test the functionality of the proposed method first, where each subject signed an informed consent.
As stated in the introduction, the experiments were also performed in three steps to illustrate the performance of different modules.\\  
\noindent {\em Step 1}: First, a motion dataset is constructed through the demonstration of three healthy subjects who wear the exoskeleton in a transparent mode \cite{zimmermann2020towards}. The collected trajectories in the dataset are used to 
train both the ProMP model (for offline generation) and the safety grader (for online adjustment). With such a probabilistic model, a new trajectory can be generated, suiting the movement habits of healthy subjects and dealing with intention uncertainties, sensor noises, and subject bias.  
\\

\noindent{\em Step 2}: 
Next, the online adaption law (\ref{reg_f}) is employed to approximate and then isolate the disturbance torque, where the parameters are set as $\alpha = 10, \bm \Gamma = diag(1,0.5,0.5)$. In this step, the robot is controlled 
to track the ProMP's mean trajectory (which satisfies the P.E. condition) without human involvement (that is, no interaction torque $\bm\tau_e=\bm 0$). 

At the steady state, the parameters of (\ref{friction_model}) converge to the values presented in Table \ref{friction_param}, such that the disturbance torque is well approximated. After compensating for these disturbances, the robot can exactly follow the reference trajectory, as shown in Fig. \ref{track}. 
A comparative study (with or without the disturbance compensation) is also shown in Fig. \ref{trackerror}, where the proposed method consistently yields smaller joint errors (average errors: $0.034~rad$ v.s. $0.068~rad$), proving the effectiveness of disturbance compensation.

\begin{table}[!h]
\centering
\renewcommand{\arraystretch}{1.2} 
\caption{Converged Friction Model}
\vspace{-0.2cm}
\label{friction_param}
    \begin{tabular}{|c|c|c|c|}
        \hline
         & $\bar a_f$ & $\bar{b}_f$ & $\bar c_f$  \\ 
         \hline
        Joint 1 & 4.014 & -1.012 & 0.311\\ \hline
        Joint 2 & 2.251 & 0.848 & -1.118\\ \hline
        Joint 4 & 6.136 & -18.632 & 16.428\\ \hline
    \end{tabular}
    \vspace{-0.3cm}
\end{table}

\begin{figure}[!h]
\vspace{-0.5cm}
\centering
\subfigure[]{
    \label{snap1}
    \includegraphics[width=1in]{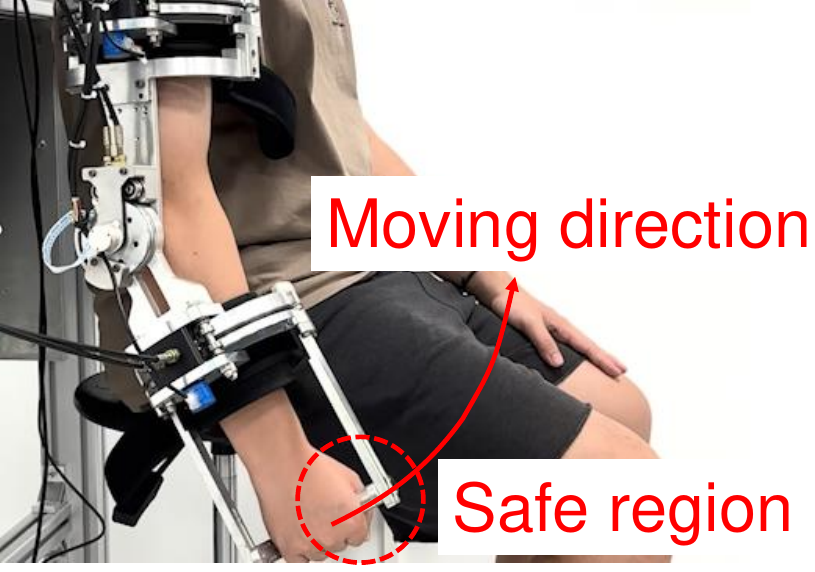}}
\subfigure[]{
    \label{snap2}
    \includegraphics[width=1in]{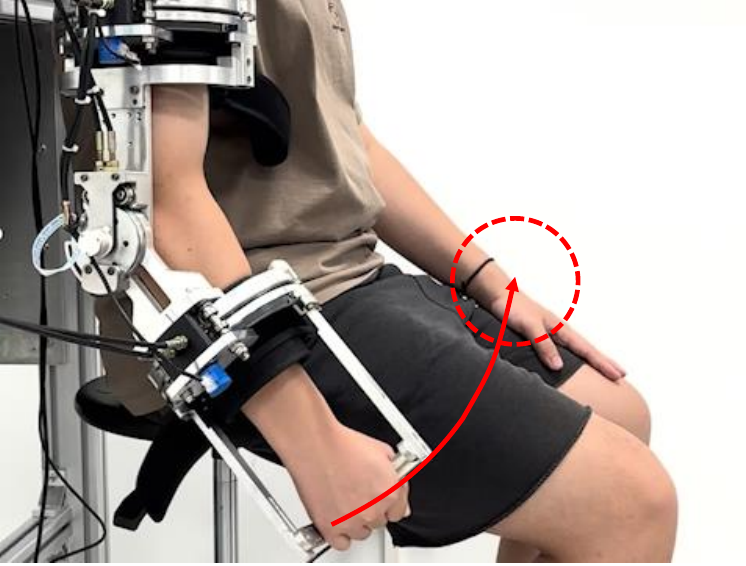}}
\subfigure[]{
    \label{snap3}
    \includegraphics[width=1in]{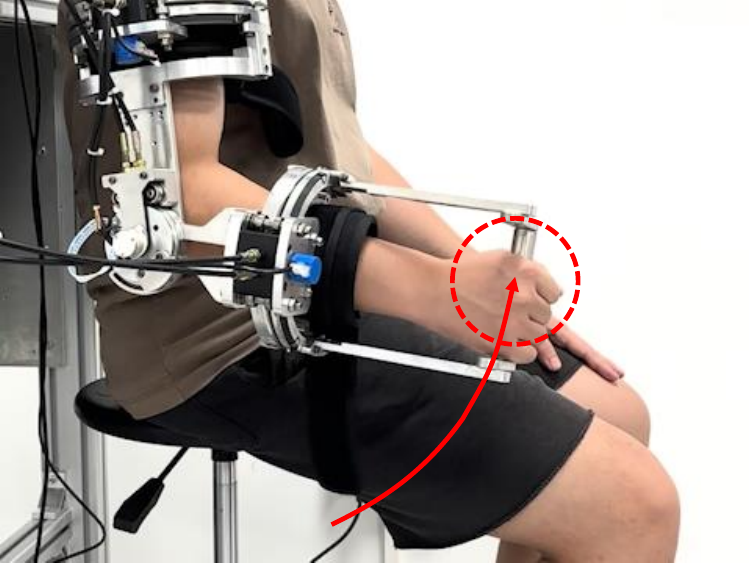}}
\subfigure[]{
    \includegraphics[width=3in]{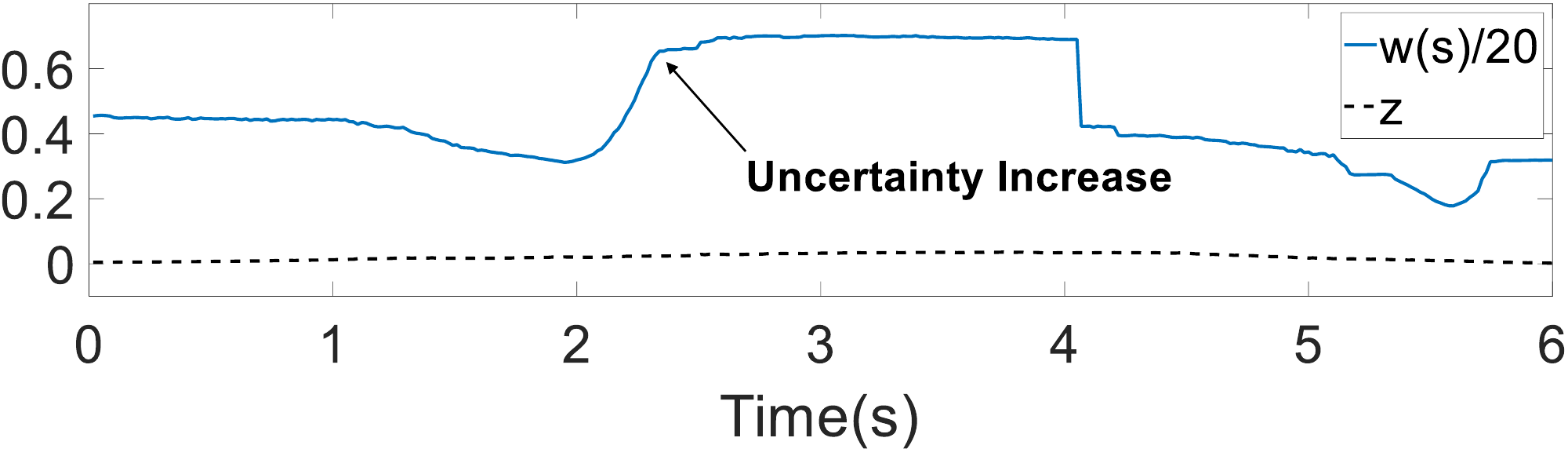}}
    \vspace{-0.3cm}
\caption{(a)-(c) Snapshots of upper-limb motion; (d) the corresponding change of the weighting function (blue solid). In the presence of varying weights, and hence varying impedance, the impedance error (black dashed) remains at zero, implying the realization of the impedance model.}\label{vary_weight}
\vspace{-0.3cm}
\end{figure}

\noindent{\em Step 3}: As mentioned, the safety grader outputs the evaluation score to form a safety region and vary the robot's impedance. Once the patient's actual motion  
leaves the safety region, the robot becomes more rigid to keep the patient away from potential hazards. 
Specifically, the parameters of the variable impedance model were set as: $\bm C_d=30\bm I_3$, $\bm K_d=50\bm I_3$, $\lambda_1=-4.5$, $m=0.1$, $h=36$, and $\lambda_2=10.5$ where $\bm I_3$ is a $3\times 3$ identity matrix. 
An example of joint motion (Joint 4) under the variable impedance model is given in Fig. \ref{vary_weight}. 
It can be seen that upon employing the developed variable impedance controller, the weighting function escalates (larger impedance) when departing from the safe region and decreases (lower impedance) as the safety issue is relaxed.

The overall performance of the proposed method is also achieved by integrating both motion planning and interaction control. Specifically, a healthy subject wears the robot and 
simulates impaired upper limb motor function by introducing additional damping to the impedance model (\ref{impedancemodel}) (e.g., intentionally holding the robot). The proposed motion planner utilizes
seven demonstrations from our database, wherein healthy subjects executed a designated rehabilitation task to initialize the ProMP. During the training process, the ProMP was also iteratively updated via physical interaction with the subject, shaping the trajectory into an increasingly individualized version. The update was stopped 
when the root mean square error (RMSE) of each joint's trajectory was less than two degrees.

\begin{figure}[!h]
\centering
\subfigure[]{
    \label{joint3}
    \includegraphics[width=0.8\linewidth]{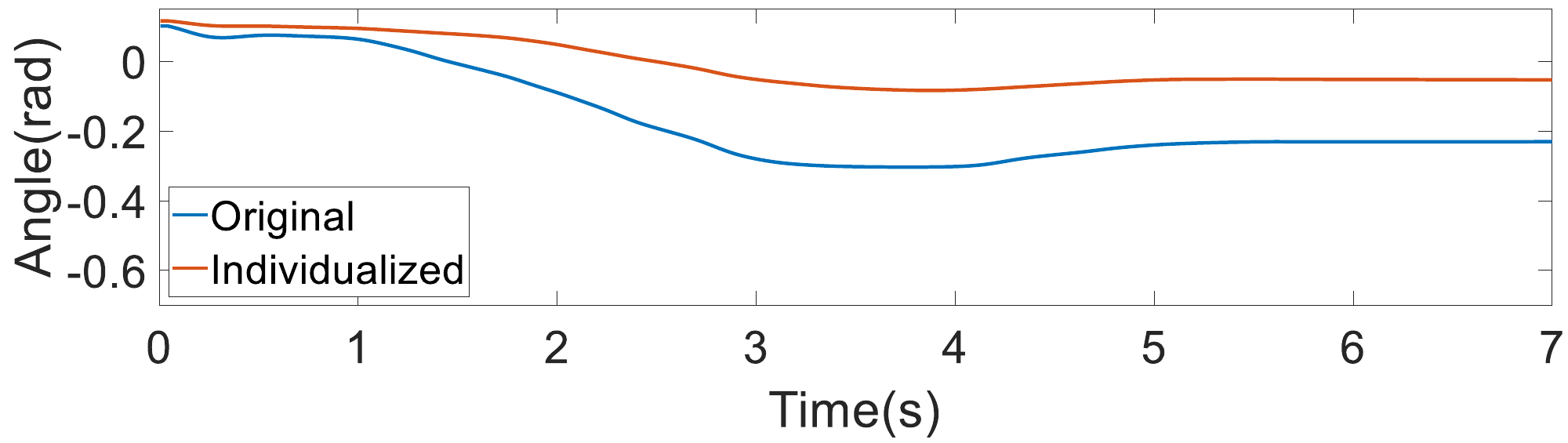}}
\subfigure[]{
    \label{joint4}
    \includegraphics[width=0.8\linewidth]{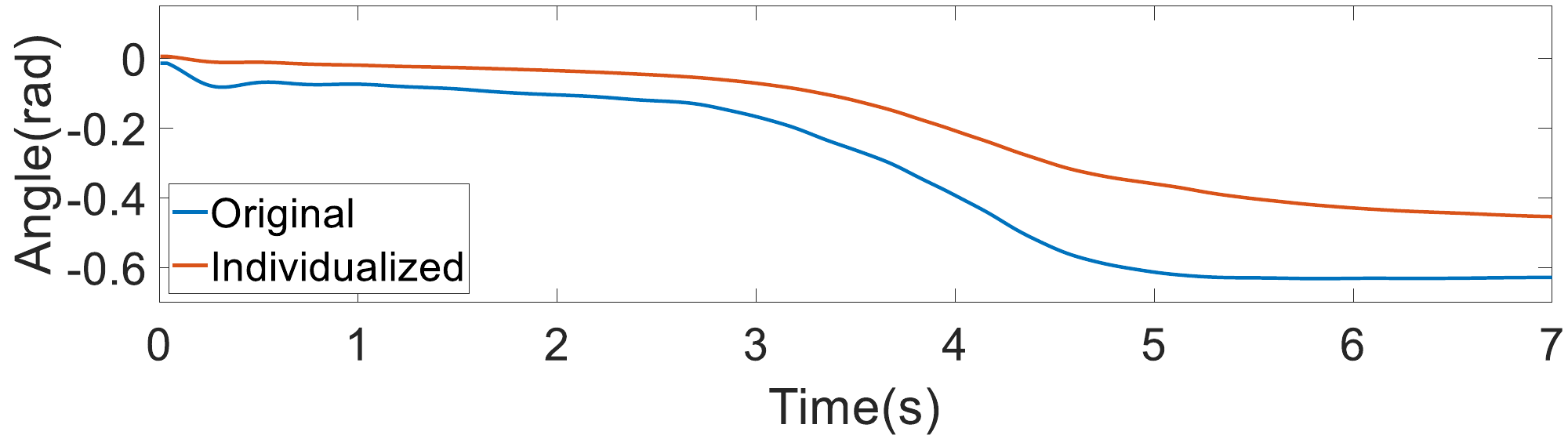}}
\subfigure[]{
    \label{joint5}
    \includegraphics[width=0.8\linewidth]{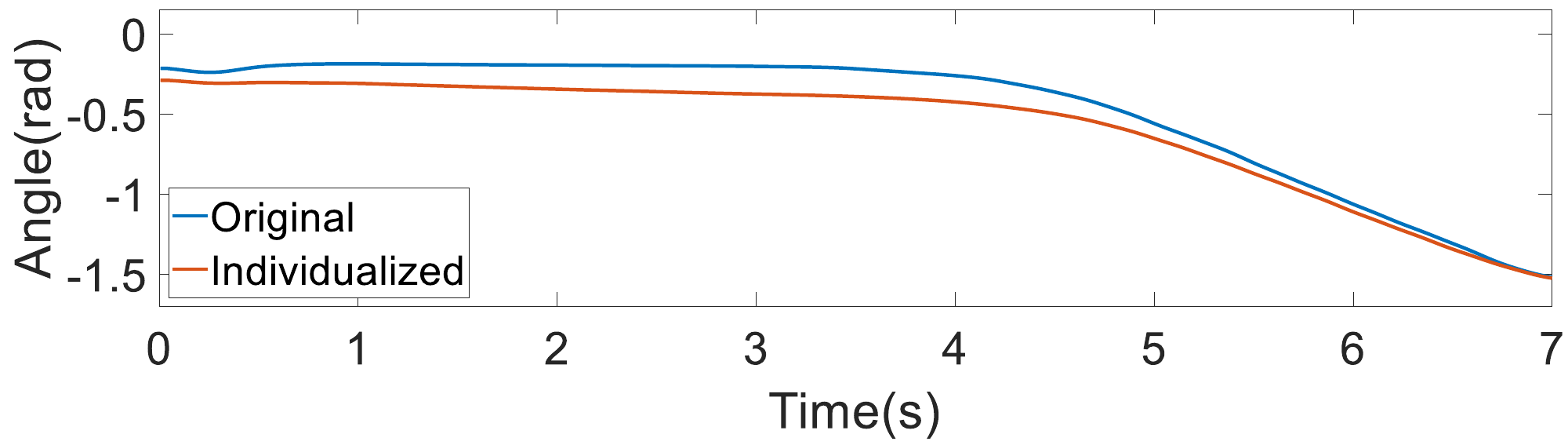}}
\caption{The comparison between the original trajectory (blue line, generated by the ProMP directly) and the individualized trajectory (red line, after online adjustment): (a) Joint 1; (b) Joint 2; (c) Joint 4.}\label{indiv_traj}
\vspace{-0.3cm}
\end{figure}

In the experimental evaluation, the planned trajectory meets our requirements after 43 demonstrations, as shown in Fig. \ref{indiv_traj}. Specifically, the individualized trajectory can better match the actual motion of the patient (compared to the original). 
Moreover, there is no significant reduction in the range of motion in the primary joint (Joint 4) during rehabilitation tasks. Otherwise, the patient might not follow the ``correct'' guidance, 
thus affecting the training effect.

\begin{figure}[!h]
    \centering
    \includegraphics[width=8cm]{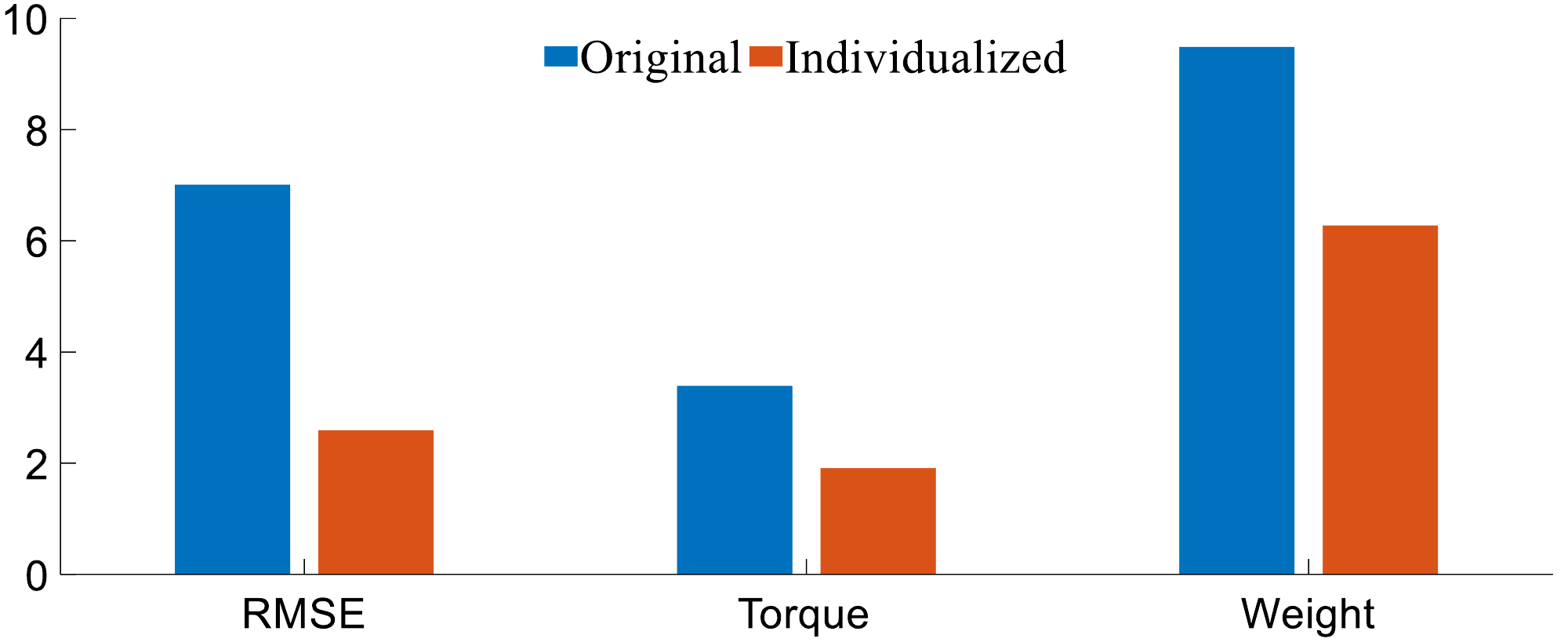}
    \caption{The difference between original and individualized assistance.
    The optimized trajectory demonstrates reduced tracking error, lower interaction torque, and a decreased weighting factor in the impedance model.}
    \label{evalua_res}
    \vspace{-0.3cm}
\end{figure}

A comparison was also carried out between the individualized and original assistance, using evaluate metrics such as RMSE ($rad$), average interaction torque ($Nm$), and the dimensionless mean impedance weighting (\ref{weight_fun}).
In the findings depicted in Fig. \ref{evalua_res}, the implementation of individualized assistance led to smaller tracking errors, lower interaction torque (i.e., less correction from the robot), and reduced weighing functions (i.e., staying inside the safe region). In other words, 
the training process was more effectively completed with less reliance on the robot (i.e., AAN). Note that the RMSE data was scaled by a factor of 100 to maintain consistency in the magnitude order.

\section{Conclusions}
In this paper, a safe and individualized motion planning scheme has been proposed for upper-limb exoskeleton robots, aiming to achieve AAN in rehabilitation training. The proposed method allows the robot to customize the assistance to suit the patient by learning from both human demonstration and interaction with the patient.  In summary, the safety of the developed robot is enhanced with the safety grader, impedance control, and compliant actuators, and its effectiveness is achieved with an individualized trajectory and accurate tracking of such a trajectory. 
The experimental results indicate improved rehabilitation task performance with reduced assistance. Future work will be devoted to expanding the motion dataset, applying it to the
active training mode, and performing clinical trials.






{\small
\bibliographystyle{ref/IEEEtran}
\bibliography{ref/IEEEabrv, ref/ref}
}

\end{document}